\documentclass[10pt,journal,compsoc]{IEEEtran}

\ifCLASSOPTIONcompsoc
  \usepackage[nocompress]{cite}
\else
  \usepackage{cite}
\fi

\ifCLASSINFOpdf
\else
\fi
\usepackage{ulem}
\usepackage{times}
\usepackage{epsfig}
\usepackage{graphicx}
\usepackage{amsmath}
\usepackage{amssymb}
\usepackage{amsmath}
\DeclareMathOperator*{\argmin}{argmin}
\usepackage[pagebackref=true,breaklinks=true,letterpaper=true,colorlinks,bookmarks=false]{hyperref}

\title{Deep-3DAligner: Unsupervised 3D Point Set Registration Network With Optimizable Latent Vector}

\author{Lingjing~Wang, ~Xiang~Li
        and~Yi~Fang
\thanks{L.Wang is with MMVC Lab, New York University Abu Dhabi, UAE and Dept. of ECE, e-mail: lingjing.wang@nyu.edu. X.Li is with the MMVC Lab, New York University Abu Dhabi, e-mail: xl1845@nyu.edu. Y.Fang is with MMVC Lab, Dept. of ECE, NYU Abu Dhabi, UAE and Dept. of ECE, NYU Tandon School of Engineering, USA, e-mail: yfang@nyu.edu.}
\thanks{Corresponding author. Email: yfang@nyu.edu}
}

\begin{document}
\maketitle

\begin{abstract}
Point cloud registration is the process of aligning a pair of point sets via searching for a geometric transformation. Unlike classical optimization-based methods, recent learning-based methods leverage the power of deep learning for registering a pair of point sets. In this paper, we propose to develop a novel model that organically integrates the optimization to learning, aiming to address the technical challenges in 3D registration. More specifically, in addition to the deep transformation decoding network, our framework introduce an optimizable deep \underline{S}patial \underline{C}orrelation \underline{R}epresentation (SCR) feature. The SCR feature and weights of the transformation decoder network are jointly updated towards the minimization of an unsupervised alignment loss. We further propose an adaptive Chamfer loss for aligning partial shapes. To verify the performance of our proposed method, we conducted extensive experiments on the ModelNet40 dataset. The results demonstrate that our method achieves significantly better performance than the previous state-of-the-art approaches in the full/partial point set registration task. 
\end{abstract}

\section{Introduction}
Point set registration is a challenging but meaningful task, which has wide application in many fields \cite{Ding_2019_CVPR,bai2007skeleton,klaus2006segment,maintz1998survey, chen2019arbicon}, such as mapping, shape recognition, correspondence, large scale scene reconstruction, and so on. Most existing non-learning methods solve the registration problem through an iterative optimization process to search the optimal geometric transformation to minimize a pre-defined alignment loss between the transformed source point set and target point set \cite{myronenko2007non,ma2013robust,ma2014robust,ling2005deformation,wang2019non}. Iterative methods usually treat registration as an independent optimization process for each given pair of source and target point sets, which cannot transfer knowledge from registering one pair to another. 

\begin{figure*}
\begin{center}
\includegraphics[width=0.8\textwidth]{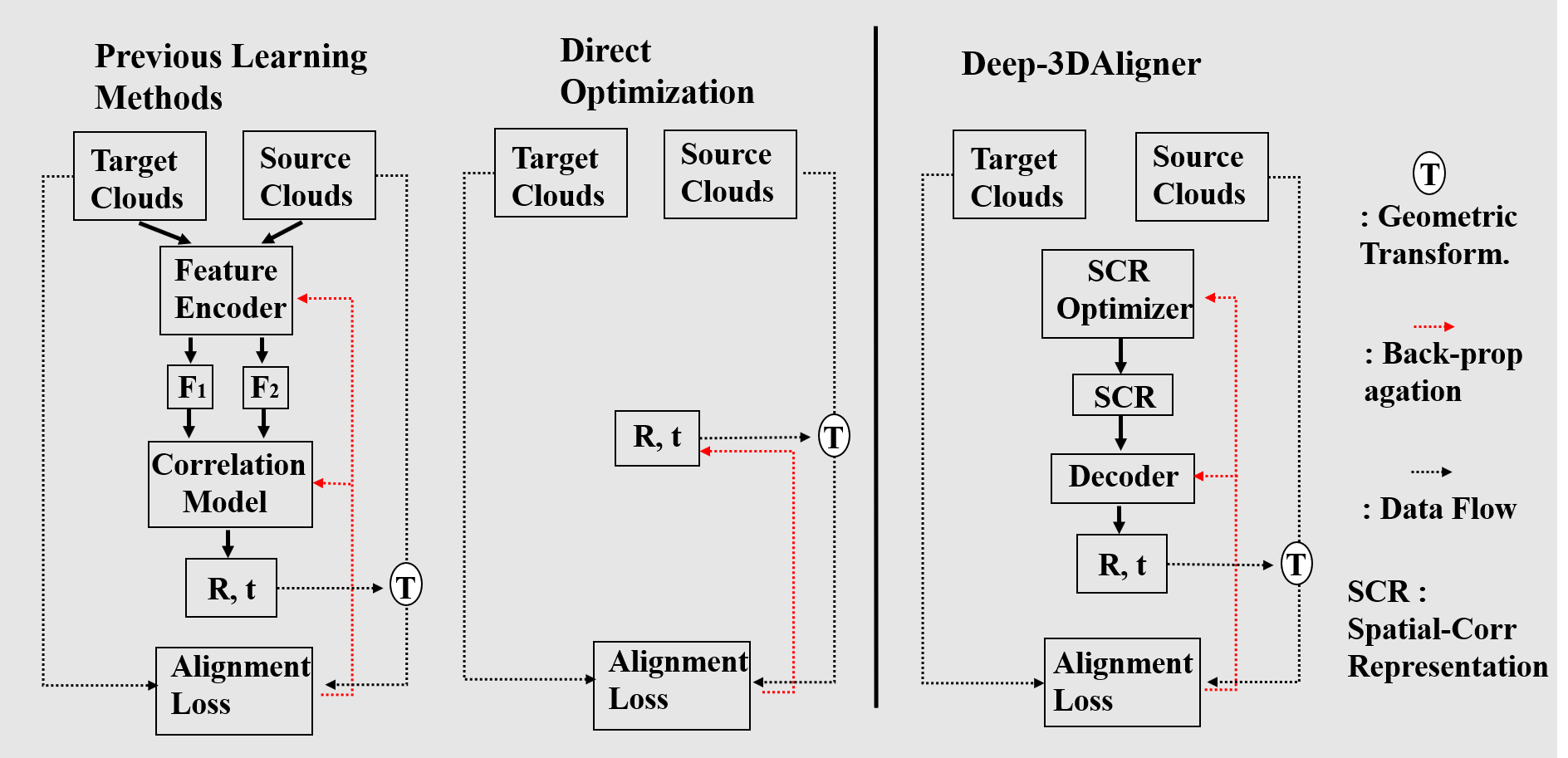}
\end{center}
\caption{Comparison of the pipeline between previous learning methods, direct optimization methods and our Deep-3DAligner for point set registration. Our method starts with optimizing a randomly initialized latent spatial correlation representation (SCR) feature, which is then decoded to the desired geometric transformation to align source and target point clouds, integrating the optimization-based SCR optimizer with the learning-based decoder to enhance the model's registration capacity.}
\label{first}
\end{figure*}

In comparison, as shown in Figure \ref{first}, instead of directly optimizing the transformation matrix towards minimization of alignment loss in non-learning based methods, learning-based methods usually leverage modern feature extraction technologies for feature learning and then regress the transformation matrix based on the mutual information and correlation defined on the extracted features of source and target shapes. The most recent model, deep closest point (DCP) \cite{wang2019deep}, leverages DGCNN \cite{wang2019dynamic} for feature learning and a pointer network to perform soft matching. Learning-based methods can greatly improve the efficiency by direct prediction of the transformation matrix for testing pairs. However, these methods' performance highly depends on the number and quality of the labeled training dataset and the performance may greatly degrade for the testing dataset of unseen categories. In contrast, our proposed network can not only leverage the deep decoding structure to learn the registration pattern from training data but also leverage a directly optimizable SCR feature to further refine the desired transformation in an unsupervised manner, which is different from the DCP that uses the ground-truth transformation parameters (i.e. rotation and translation matrix) for training.

For most cases in practice, input point sets may suffer from various noise such as shape incompleteness \cite{wang2019prnet}. When dealing with partial shapes, classical deep learning-based methods suffer from performance degradation, especially when the overlapping subsets between source and target shapes are small. Wang et al. \cite{balakrishnan2018unsupervised} proposed the first registration method designed for solving the partial point set registration problem. This method still requires the process of detecting key corresponding points from the partial input shapes. When the overlapping area is small, the corresponding points are difficult to be detected, which can lead to inferior registration performance. In contrast, we propose an adaptive Chamfer loss to detect the corresponding subsets between source and target point sets instead of a few key points. Driven by this designed loss, the desired geometric transformation can be gradually optimized based on the detected overlapping subsets between source and target point sets in a coarse-to-fine way.

With the development of the SCR feature, our proposed Deep-3DAligner framework is illustrated in Figure \ref{main}, which contains three main components. The first component is an SCR optimizer where the deep SCR feature is optimized from a randomly initialized feature. The second component is a transformation decoder which decodes the SCR feature to regress the transformation parameters for the point sets alignment. The third component is an alignment loss that measures the similarity between the transformed source point set and the target one. In the pipeline, there are two communication routes, indicated by black and red dashed lines. The communication route in black is for the data flow for the Deep-3DAligner paradigm, where the source and target point sets are used as input. The communication route in red is the back-propagation route with which the alignment loss is back-propagated to update the SCR and the transformation decoder. Our contribution is as follows:
\begin{itemize}
\item We introduce a novel unsupervised learning approach for the point set registration task.
\item We introduce a spatial correlation representation (SCR) feature which can eliminate the design challenges for encoding the spatial correlation between source and target point sets in comparison to learning-based methods. 
\item We propose an adaptive Chamfer loss to gradually detect the overlapping areas between the transformed source point set and target point set to refine the desired geometric transformation in a coarse-to-fine approach.
\item Experimental results demonstrate the effectiveness of the proposed method for point set registration, and even without ground truth transformation for training, our proposed approach achieved superior performance in 3D full/partial point sets registration compared to most recent supervised state-of-the-art approaches.
\end{itemize}

\section{Related Works}
\subsection{Iterative registration methods}
The development of optimization algorithms to estimate rigid and non-rigid geometric transformations in an iterative routine has attracted extensive research attention in past decades. The iterative closest point (ICP) algorithm \cite{besl1992method} is one successful solution for rigid registration. It initializes an estimation of a rigid function and then iteratively chooses corresponding points to refine the transformation. Go-ICP \cite{yang2015go} was further proposed by Yang et al. to leverage the BnB scheme for searching the entire 3D motion space to solve the local initialization problem brought by ICP. Zhou et al. proposed fast global registration \cite{zhou2016fast} for the registration of partially overlapping 3D surfaces. The TPS-RSM algorithm was proposed by Chui and Rangarajan \cite{chui2000new} to estimate parameters of non-rigid transformations with a penalty on second-order derivatives. Coherence point drift (CPD) was further proposed by Myronenko et al. \cite{myronenko2007non} for non-rigid point set registration. Although the independent iterative optimization process limits the efficiency of registering a large number of pairs, inspiring us to leverage its advantage and equip it with a learning-based system for this task.

\subsection{Learning-based registration methods} Recent works have started a trend of directly learning geometric features from cloud points (especially 3D points), which motivates us to approach the point set registration problem using deep neural networks \cite{balakrishnan2018unsupervised,zeng20173dmatch,qi2017pointnet}. PointNetLK \cite{aoki2019pointnetlk} was proposed by Aoki et al. to leverage the newly proposed PointNet algorithm for directly extracting features from the point cloud with the classical Lucas $\&$ Kanade algorithm for the rigid registration of 3D point sets. Liu et al. proposed FlowNet3D \cite{liu2019flownet3d} to treat 3D point cloud registration as a motion process between points. Wang et al. proposed a deep closest point \cite{wang2019deep} model, which first leverages the DGCNN structure to exact the features from point sets and then regress the desired transformation based on it. Balakrishnan et al. \cite{balakrishnan2018unsupervised} proposed a voxelMorph CNN architecture to learn the registration field to align two volumetric medical images. In contrast, we first propose a model-free structure to skip the encoding step. Instead, we initialize an SCR feature without pre-defining a model, which is to be optimized with the weights of the network from the alignment loss back-propagation process. PR-Net \cite{wang2019prnet} as the first work proposed a method to detect corresponding points from partial shapes and then solve the desired geometric transformation based on it. In comparison, our method detects the overlapping subsets based on our proposed adaptive Chamfer loss.

\begin{figure*}
\begin{center}
\includegraphics[width=0.8\textwidth]{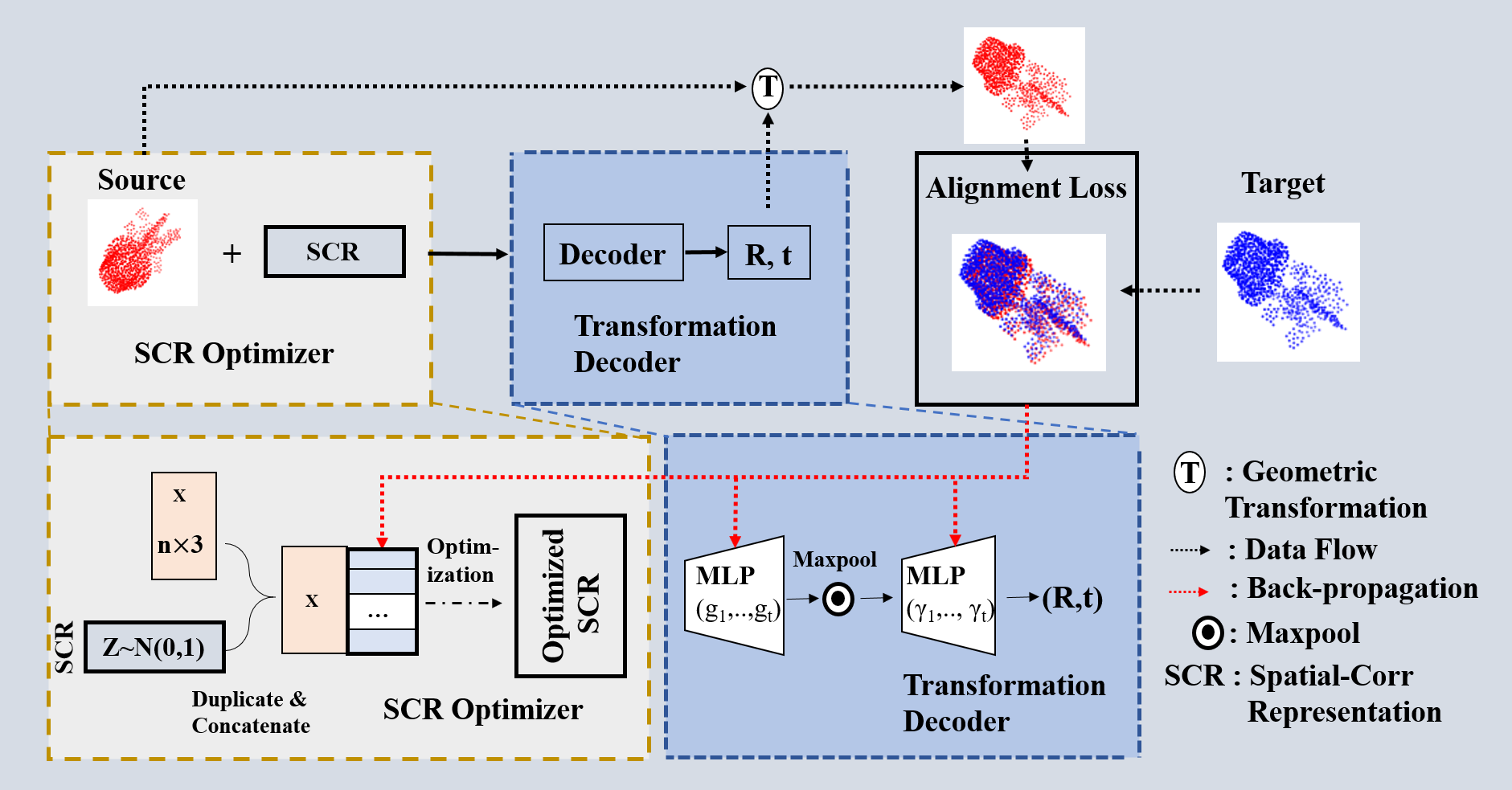}
\end{center}
\caption{Our pipeline. For a pair of input source and target point sets, our method starts with the SCR optimization process to generate a spatial-correlation representation feature, and a transformation regression process further decodes the SCR feature to the desired geometric transformation. The alignment loss is back-propagated to update the weight of the transformation decoder and the SCR feature during the training process. For testing, the weights of the transformation decoder remain constantly without updating.}
\label{main}
\end{figure*}

\section{Approach}
We introduce our approach in the following sections. First, we define the learning-based registration problem in section \nameref{sc_problem}. In section \nameref{sc_scr}, we introduce our spatial-correlation representation. The transformation decoder is illustrated in section \nameref{sc_trans_reg}. In section \nameref{sc_loss}, we provide the definition of the loss function. Section \nameref{sc_optim} illustrates the newly defined optimization strategy. 

\subsection{Problem statement} \label{sc_problem}
Given training dataset $\mathbf{D}=\{(S_i, G_i) \text{ ,where } S_i \in \mathbf{S}, G_i \in \mathbf{G}\}$. $\mathbf{S}$ is the source point set and $\mathbf{G}$ is the target point set, and we have $\mathbf{S},\mathbf{G} \subset \mathbb{R}^N (N=2 \text{ or } N=3)$. Assuming the existence of a function $g_{\theta}(S_i,G_i) = \phi$ and $g$ has parameter set $\theta$ in a neural network structure. When considering only rigid point set registration, the output $\phi$ usually contains a homogeneous transformation matrix including a rotation matrix and a translation vector. A model with optimized weights $\theta^{optimal}$ can generate the desired rotation and translation parameters $\phi$ in $g$ to further align the source and target point sets. The objective loss function $\mathcal{L}$ is usually a pre-defined similarity metric for evaluation of the alignment quality between transformed source and target point sets. Based on a given dataset $\mathbf{D}$, a stochastic gradient-descent-based algorithm can be used to update the weight parameters $\theta$ and to minimize the pre-defined loss function:
\begin{equation}
\begin{split}
\theta^{optimal} =\argmin_{\theta}[\mathbb{E}_{(S_i,G_i)\sim \mathbf{D}}[\mathcal{L}(S_i,G_i, g_{\theta}(S_i,G_i))]]
\end{split}
\end{equation}

\subsection{Spatial-Correlation Representation}\label{sc_scr}
In this paper, we define the spatial correlation representation as the latent feature that characterizes the essence of spatial correlation between a given pair of source and target point sets. As shown in Figure \ref{first}, to compute the SCR feature, source and target point sets are usually fed to a feature in previous works for the deep spatial feature extraction, and followed with a pre-defined correlation module. However, the design of an appropriate feature encoder for unstructured point clouds is challenging compared to the standard discrete convolutions assume the availability of a grid structured input (e.g. 2D image). The limitation of the hand-crafted design of modules for the extraction of individual spatial feature and spatial correlation feature motivates us to design a model-free based SCR as described below. 

To eliminate the side effects of the hand-craft design in feature encoder and correlation module and to better equip the optimization process within the system, as shown in Figure \ref{main}, we define a trainable latent SCR (Spatial-Correlation Representation) feature for each pair of point sets. The design of SCR makes it possible to both leverage the common ``knowledge'' of registration from the dataset via the shared generator and individual adjustment for each input pair through the optimization of their SCR. After optimization, SCR should contain the spatial correlation information for each input pair. As shown in Figure \ref{main}, for a pair of source and target point sets  $S_i$ and $G_i$, the randomly initialized latent vector $z_i \sim \mathcal{N}(0,0.01)$ from Gaussian distribution as an initialized SCR. The initialized SCR is optimized during the training process together with the transformation decoder. The implicit design of SCR allows Deep-3DAligner more flexibility in spatial correlation feature learning that is more adaptive for the alignment of unseen point sets and partial point sets as well.  

\subsection{Transformation Decoder}\label{sc_trans_reg}
Given the above spatial-correlation representation (SCR) feature, we then design a decoding network to regress the desired transformation parameters, as illustrated in Figure \ref{main}. More specifically, $\forall x \in S_i$, we stack the coordinates of $x$ with $z_i$ and we note it as $[x, z_i]$. We leverage a multi-layer perceptron (MLP) structure to regress the rotation and translation parameters in the desired transformation. For each layer of the MLP, we define $\{g_i\}_{i=1,2,...,s}$, such that $g_i : \mathbb{R}^{v_{i}}\to \mathbb{R}^{v_{i+1}}$, where $v_{i}$ is the dimension of input layer and $v_{i+1}$ is the dimension of output layer. For each MLP layer, we use the ReLU activation function. For the output of last layer, we leverage a max pool function to accumulate the point features into a latent vector $L_i$, calculated as:

\begin{equation}
\begin{split}
L_i=Maxpool\{g_sg_{s-1}...g_1([x_j,z_i])\}_{x_j \in S_i}
\end{split}
\end{equation}

Taking the latent vector $L_i$ as the input, we have a further $t$ successive MLP layers with a ReLU activation function to regress the parameters $\phi_i$ of the desired transformation. We define function $\{\gamma_i\}_{i=1,2,...,t}$, such that $\gamma_i : \mathbb{R}^{w_{i}}\to \mathbb{R}^{w_{i+1}}$, where $w_{i}$ is the dimension of input layer and $w_{i+1}$ is the dimension of output layer. 
\begin{equation}
\begin{split}
\phi_i=\gamma_t\gamma_{t-1}...\gamma_1(L_i)
\end{split}
\end{equation}

We further compute the transformed source point set, defined as $S_i'$,

\begin{equation}
\begin{split}
S_i' =\mathbf{\mathbf{T}_{\phi_i}}(S_i)
\end{split}
\end{equation}
where $\mathbf{T}_{\phi_i}$ denotes the desired geometric transformation with parameters $\phi_i$. Based on the transformed source point set $S_i'$ and the target point set $G_i$, we further introduce the loss function. 

\subsection{Loss function}\label{sc_loss}
In our unsupervised setting, we do not have the ground truth transformation for supervision and we do not assume a direct correspondence between these two point sets. Therefore, a distance metric between two point sets, instead of the point/pixel-wise loss is desired. In addition, A suitable metric should be differentiable and efficient to compute. In this paper, we adopt the Chamfer distance as our loss function. The Chamfer loss is a simple and effective alignment metric defined on two non-corresponding point sets. We formulate the Chamfer loss between our transformed source point set $S_i'=T_{\phi}(S_i)$ and target points set $G_i $ as:
\begin{equation} 
\begin{split}L_{\text{Chamfer}}(S_i',G_i)
 &= \sum_{x\in S_i'}\min_{y \in G_i}||x-y||^2_2\\
 &+ \sum_{y\in G_i}\min_{x \in S_i'}||x-y||^2_2
\end{split}
\end{equation}

For aligning partial-shapes, we further propose an adaptive Chamfer loss. For a time period $t$ during the optimization process, we assume a pre-defined distance threshold $\sigma_t$. For the transformed source point set $S_i'$, we define the overlapping subset of it with the target point set $G_i$ as ${S'_i}^{(t)} \subset {S'_i}^{(t-1)} \subset ... \subset {S'_i}^{(0)}=S_i'$, such that $\forall x \in {S'_i}^{(t-1)}, \text{if \ } \min\{||x-y||^2_2\}_{y \in G_i^{(t-1)}} < \sigma_t$, then $x \in {S'_i}^{(t)}$. Similarly, for the target point set $G_i$, we define the overlapping subset between it with $S_i'$ as ${G_i}^{(t)} \subset {G_i}^{(t-1)} \subset ... \subset {G_i}^{(0)}=G_i$, such that $\forall x \in {G_i}^{(t-1)}, \text{if \ } \min\{||x-y||^2_2\}_{y \in {S'_i}^{(t-1)}} < \sigma_t$, then $x \in {G_i}^{(t)}$. Therefore, based on the overlapping subsets ${S_i'}^{(t)}$ and ${G_i}^{(t)}$ for time period $t$, we define the adaptive Chamfer loss for $S_i'$ and $G_i$ as:
\begin{equation} 
\begin{split}L^t_{\text{Adaptive-Chamfer}}(S_i',G_i)
&= \sum_{x\in S_i'^{(t)}}\min_{y \in G_i^{(t)}}||x-y||^2_2\\
&+ \sum_{y\in G_i^{(t)}}\min_{x \in S_i'^{(t)}}||x-y||^2_2 
\end{split}
\end{equation}

\subsection{Optimization Strategy}\label{sc_optim}

In section \nameref{sc_scr}, we define a set of trainable latent vectors $\mathbf{z}$, one for each pair of point sets as the SCR feature. During the training process, these latent vectors are optimized along with the weights of network decoder using a stochastic gradient descent-based algorithm. For a given training dataset $\mathbf{D}$, our training process can be expressed as:
\begin{equation}
\begin{split}
\mathbf{\theta^{optimal}, z^{optimal}} =\argmin_{\theta, \mathbf{z}}[\mathbb{E}_{(S_i,G_i)\sim \mathbf{D}}[\mathcal{L}(S_i,G_i, g_{\theta}(S_i,z_i))]],
\end{split}
\end{equation}
where $\mathcal{L}$ represents the pre-defined loss function.

For a given testing dataset $\mathbf{W}$, we fix the network parameters $\tilde{\theta}= \mathbf{\theta^{optimal}}$ and only optimize the SRC features:
\begin{equation}
\begin{split}
\mathbf{z^{optimal}} =\argmin_{\mathbf{z}}[\mathbb{E}_{(S_j,G_j)\sim \mathbf{W}}[\mathcal{L}(S_j,G_j, g_{\tilde{\theta}}(S_j,z_j))]].
\end{split}
\end{equation}
The learned decoder network parameters $\tilde{\theta}$ here provides a prior knowledge for the optimization of SRC. After this optimization process, the desired transformation can be determined by $T_{\phi_i}=T_{g_{\tilde{\theta}}(S_i,z_i^{optimal})}$ and the transformed source shape can be generated by $S_i'=T_{\phi_i}(S_i),  \forall S_i \in \mathbf{W}$. 

\section{Experiments}\label{sc_exp}
\subsection{Dataset}\label{sc_dataset}
We test the performance of our model for 3D point set registration on the ModelNet40 dataset. This dataset contains 12311 pre-processed CAD models from 40 categories. For each 3D point object, we uniformly sample 1024 points from its surface. Following the settings of previous work, points are centered and re-scaled to fit in the unit sphere. For each source shape $S_i$ we generate the transformed shapes $G_i$ by applying a rigid transformation defined by the rotation matrix which is characterized by 3 rotation angles along the x-y-z-axis, where each value is uniformly sampled from $[0,45]$ unit degree, and the translation which is uniformly sampled from $[-0.5, 0.5]$. At last, we simulate partial point sets by randomly select a point in unit space and keep its 768 nearest neighbors for source and target shapes.

\begin{figure*}
\begin{center}
\includegraphics[width=0.8\textwidth]{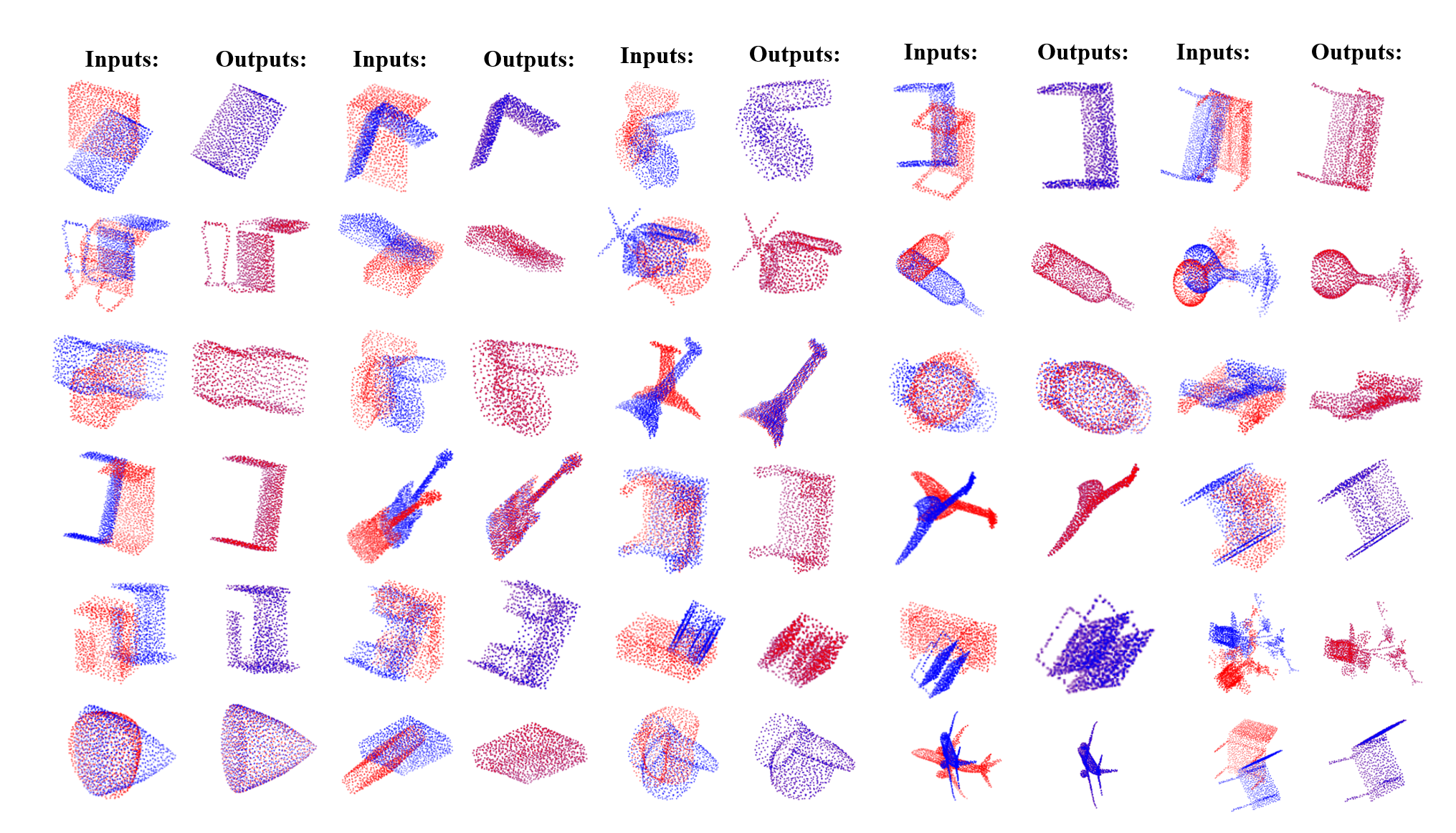}
\end{center}
\caption{Randomly selected qualitative results of our model for registration of unseen samples. Left columns: inputs. Right columns: outputs. The red points represent source point sets, and the blue points represent the target point sets.}
\label{allres}
\end{figure*}

\begin{table*}
\small
\begin{center}
 \resizebox{\textwidth}{!}{\begin{tabular}{ccccccc}
\hline
Model& MSE(R) &RMSE(R)& MAE(R) &MSE(t)& RMSE(t)& MAE(t)
\\
 \hline
Direct Optimization &406.131713&16.454065&13.932246&0.087263&0.295404&0.253658\\
ICP \cite{besl1992method}& 894.897339& 29.914835& 23.544817& 0.084643 &0.290935& 0.248755\\
Go-ICP \cite{yang2015go}& 140.477325& 11.852313& 2.588463& 0.000659& 0.025665& 0.007092\\
FGR \cite{zhou2016fast}& 87.661491& 9.362772& 1.999290& 0.000194& 0.013939 &0.002839\\
PointNetLK \cite{aoki2019pointnetlk}& 227.870331& 15.095374& 4.225304& 0.000487& 0.022065& 0.005404\\
DCPv1+MLP(Supervised)\cite{wang2019deep} & 21.115917& 4.595206&  3.291298& 0.000861 &0.029343&0.022501\\
DCPv2+MLP(Supervised)\cite{wang2019deep} & 9.923701&  3.150191& 2.007210&0.000025&  0.005039& 0.003703\\
\hline
DCPv1+SVD(Supervised)\cite{wang2019deep} & 6.480572& 2.545697& 1.505548& 0.000003 &0.001763& 0.001451\\
DCPv2+SVD(Supervised)\cite{wang2019deep} & 1.307329& 1.143385&0.770573& \textbf{0.000003}& \textbf{0.001786}& \textbf{0.001195}\\
\hline
Ours (MLP-based, Unsupervised) & \textbf{0.220650} &\textbf{0.350199}&\textbf{0.248512}&0.000149 &0.008881& 0.005021\\
\hline
\end{tabular}}
\end{center}
\caption{ModelNet40: Test on unseen point clouds. Our model is trained in an unsupervised manner without  any ground-truth labels. Our model does not require attention mechanism and SVD-based fine-tuning processes.}
\label{ttt2}
\end{table*}

\begin{table*}
\small
\begin{center}
 \resizebox{\textwidth}{!}{\begin{tabular}{ccccccc}
\hline
Model& MSE(R) &RMSE(R)& MAE(R) &MSE(t)& RMSE(t)& MAE(t)
\\
 \hline
ICP\cite{besl1992method}& 892.601135&29.876431&23.626110&0.086005&0.293266&0.251916\\
Go-ICP \cite{yang2015go}& 192.258636&13.865736&2.914169&0.000491&0.022154&0.006219\\
FGR \cite{zhou2016fast}&97.002747&9.848997&1.445460&0.000182&0.013503&0.002231\\
PointNetLK \cite{aoki2019pointnetlk}& 306.323975&17.502113&5.280545&0.000784&0.028007&0.007203\\
DCPv1+SVD (Supervised) \cite{wang2019deep} &  19.201385&4.381938&2.680408&0.000025&0.004950&0.003597\\
DCPv2+SVD (Supervised) \cite{wang2019deep} & 9.923701& 3.150191& 2.007210& \textbf{0.000025} &\textbf{0.005039}&\textbf{0.003703}\\
Ours (MLP-based, Unsupervised) & \textbf{0.280846} &\textbf{0.398275}&\textbf{0.287559}&0.000088 &0.007547& 0.004629\\
\hline
\end{tabular}}
\end{center}
\caption{ModelNet40: Test on unseen categories. Our model is trained in an unsupervised manner without ground-truth labels. Our model does not require SVD-based fine-tuning processes.}
\label{ttt5}
\end{table*}

\begin{figure*}
\centering
\includegraphics[width=0.8\textwidth]{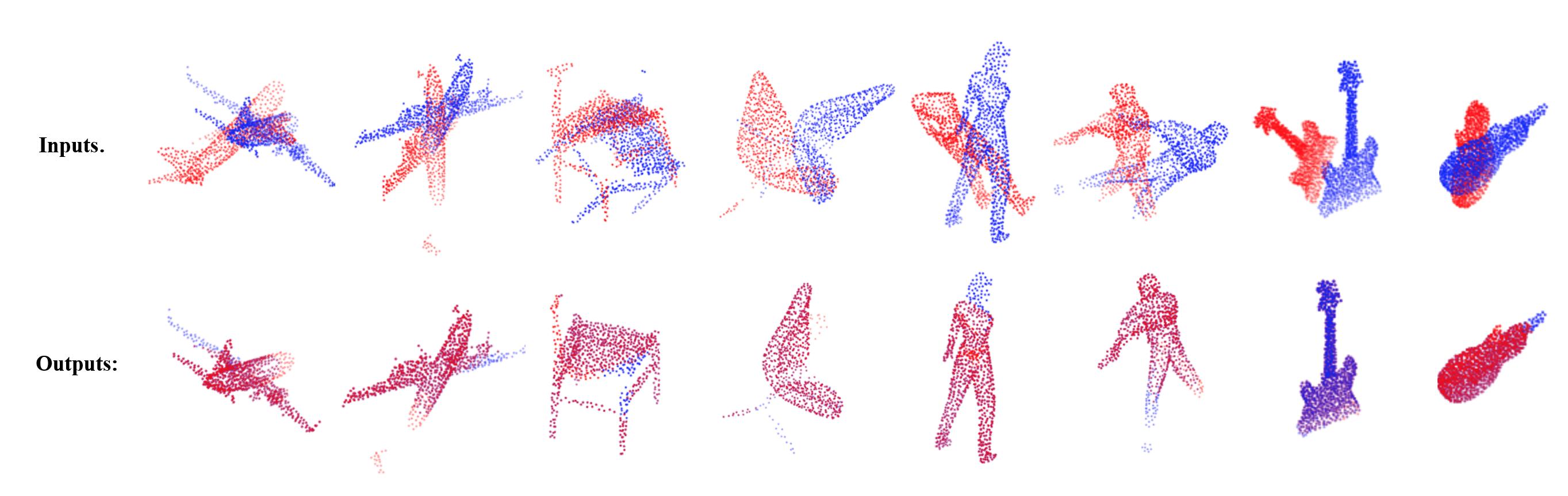}
\caption{Qualitative results of partial shapes alignment. From left to right: selected results on the airplane,chair, human, guitar category respectively.}
\label{fig:4}
\end{figure*}

\subsection{Settings} \label{sc_setting}
We train our network using batch data from the training data set $\{(S_i,G_i) | S_i, G_i \in \mathbf{D} \}_{i=1,2,...,b}$. We set the batch size $b$ to 128. The latent vectors are initialized from a Gaussian distribution $\mathcal{N}(0,0.01)$ with a dimension of 256. For the Deep-3Daligner network, the first part includes 2 MLP layers with dimensions (256,128) and a max pool layer. Then, we use 3 additional MLPs with dimensions of (128, 64, 3) for decoding the rotation matrix and with dimensions of (128, 64, 3) for decoding the translation matrix. We use the leaky-ReLU \cite{xu2015empirical} activation function and implement batch normalization \cite{ioffe2015batch} for every layer except the output layer. For adaptive Chamfer, $\sigma_t$ decreases from 10 to 0.01 in 100 epochs. The learning rate is set as 0.001 with exponential decay of 0.995 at each epoch. We use the mean squared error (MSE), root mean squared error (RMSE), and mean absolute error (MAE) to measure the performance of our model and all comparing methods. Lower values indicate better alignment performance. All angular measurements in our results are in units of degrees. The ground-truth labels are only used for the performance evaluation and are not used during the training/testing process.

\subsection{Full 3D point set registration}
In this experiment, we follow previous works to test our model for 3D point set registration on unseen point sets in Test 1, and on unseen categories in Test 2. \\

\noindent{\textbf{Experiment Setting:}} In Test 1, for the 12,311 CAD models from the ModelNet40, following exactly DCP's setting, we split the dataset into 9,843 models for training and 2,468 models for testing. In Test 2, to test the generalizability of our model, we split ModelNet40 evenly by category into training and testing sets in the same way as DCP. We train our Deep-3DAligner, DCP, and PointNetLK on the divided training dataset and then evaluate the performance on the testing set. ICP, Go-ICP, and FGR are tested directly on the testing dataset. Note that our model is trained without using any ground-truth information and our model does not require the SVD-based fine-tuning processes.\\

\noindent{\textbf{Results of Test 1 (the training/testing split test): }} We list the quantitative experimental results in Table \ref{ttt2}. In this table, we evaluate the performance based on the prediction errors of rotation angles and translation vectors. The first three columns illustrate the comparison results for the rotation angle prediction. For reporting this performance, we ignore the categories of exact symmetric shapes' quantitative results for rotation matrix since in theory there is no unique solution for these cases. For example, for the first cone shown in row 6 of Figure \ref{allres}, even though the alignment is perfect, we cannot find the unique desired rotation matrix. These categories include bottle, bowl, cone, cup, vase. As shown in Figure \ref{allres}, we randomly select the qualitative results from the testing dataset. As we can see from the results, our method achieves significantly better performance than the baseline DCPv1+MLP model and also get even better or comparative performance against the DCPv2+SVD version even though we do not require label information for training and we do not require additional SVD layer for fine-tuning. Moreover, considering that DCP assumes the same sampling of points for source and target shapes, we tested that DCP experienced a severe performance degradation in MSE(t) for randomly sampled points of source and target shapes, whereas our model with Chamfer distance is robust to the way of point sampling. For an additional oblation study, we list the performance of direct optimization described in Figure \ref{first}. We notice that the performance of direct optimization without a learning-based decoder is not satisfied.  \\


\begin{table*}
    \centering
    \small
    \begin{tabular}{lrrrrrr}
        \hline
        Model & MSE(R) & RMSE(R) & MAE(R) & MSE(t) & RMSE(t) & MAE(t) \\
        \hline
        DCP \cite{wang2018deep} & 17.078770&4.132647&3.095657&0.001602&0.040024&0.029227\\
        PR-Net \cite{wang2019prnet} &9.120225&3.019970&1.371537&0.000286&0.016917&0.011078\\
        Ours&\textbf{0.001736} &\textbf{0.041205}&\textbf{0.030796}&\textbf{0.00000003}& \textbf{0.000174}&\textbf{0.000134}\\
        \hline
        \hline
       DCP \cite{wang2018deep} & 26.890577&5.185613&3.817362&0.001259&0.035477&0.026415\\
        PR-Net \cite{wang2019prnet} &9.430604&3.070928&1.388999&0.000296&0.017214&0.011231\\
        Ours &\textbf{0.011774}&\textbf{0.107868}&\textbf{0.064698}&\textbf{0.000056}&\textbf{0.007541}&\textbf{0.002581}\\
        \hline
        \hline
        DCP \cite{wang2018deep} & 26.444530&5.142425&3.424549&0.003018&0.054939&0.039508\\
        PR-Net \cite{wang2019prnet} & 15.00834&3.874060&1.41152&0.000296&0.017232&0.011141\\
        Ours & \textbf{0.021988}& \textbf{0.143298}& \textbf{0.086875}&\textbf{0.000021}& \textbf{0.004583}& \textbf{0.001227}\\
        
        \hline
        \hline
        DCP \cite{wang2018deep} & 34.579647&5.880446&4.426307&0.001672&0.040888&0.031120\\
        PR-Net \cite{wang2019prnet} & 8.569474&2.927366&1.368731&0.000291&0.017074&0.011149 \\
        Ours & \textbf{0.026013}& \textbf{0.157938}& \textbf{0.075724}& \textbf{0.000029}& \textbf{0.005431}& \textbf{0.001823}\\
        \hline
    \end{tabular}
    \caption{Testing performance on partial shapes alignment. From top to bottom: test performance on the chair, airplane, human, guitar category respectively.}
    \label{table2}
\end{table*}

\noindent{\textbf{Results of Test 2 (seen/unseen categories test): }} As shown in Table \ref{ttt5}, the quantitative results indicate that our model achieves superior generalization ability on unseen categories as an unsupervised method. For reporting this performance, we ignore the categories of exact symmetric shapes' quantitative results for rotation matrix since in theory there is no unique solution for these cases as explained in the previous part for Test 1. In comparison, all the supervised learning methods experienced a dramatic performance drop compared to the results in Table \ref{ttt2}. For example, we can see that PointNetLK and DCPv2+SVD obtain an MSE(R) of 227.87 and 1.31 in ``the training/testing split test''  as described in the previous experiment (see Table \ref{ttt2}). However, the corresponding values in ``seen/unseen categories test'' as described in this section increase to 306.32 and 9.92 respectively (see Table \ref{ttt5}). The MSE(R) of PointNetLK increased from 227.87 for unseen point clouds to 306.324 for unseen categories. Unsupervised algorithms, such as ICP and FGR, achieve similar accuracy for unseen categories and unseen point clouds. Our method has a small performance drop for the unseen categories compared to the results for unseen point clouds. Particularly, in the prediction of the rotation matrix for unseen categories, our method outperforms state-of-the-art DCPv2-SVD by a large margin (3400\% improvement) in MSE(R).

\subsection{Partial 3D point set registration}
In this experiment, we further verify the performance of our model for registering partial shapes.\\

\noindent{\textbf{Experiment Setting:}} For this experiment, we use four categories (chair, human, guitar, and airplane) from the ModelNet40 dataset to compare the performance of our model with state-of-the-art methods. In this section, we individually adjust and test our model for each category by controlling the learning rate with the threshold $\sigma_t$ described in adaptive chamfer loss.  For a fair comparison, we follow the settings of PR-Net paper to keep consistency with the previous methods as explained in the setting part of the experiment section. We compare our model with DCP and PR-Net in this section. Furthermore, we show the registration time for registering 100 pairs of 3D shapes between classical iterative method ICP, learning-based methods PR-net and DCP, and our hybrid method. We run DCP, PR-Net, and our model using a single 12-GB Tesla K80 GPU and ICP using CPU.\\

\begin{table}[ht]
\small
\begin{center}
{\begin{tabular}{ccccc}
\hline
Models& ICP &DCP& PR-Net &Ours\\
\hline
Time& 571s &4s&5s&66s \\
\hline
\end{tabular}}
\end{center}
\caption{Running time for aligning 100 pairs of 3D point clouds from test dataset.}
\label{ttttt}
\end{table}

\noindent{\textbf{Results:}}  We list the quantitative experimental results in Table \ref{table2}. As we can see from the results, our method achieves significantly better performance than PR-Net and DCP models regarding the results of rotation angle and translation prediction for all four categories. Since we individually adjust our model for each category, our model can achieve superior performance in comparison to DCP and PR-Net. Regarding the running time, as shown in Table \ref{ttttt}, for aligning 100 shapes from the testing dataset, our model spent 66 seconds. We do sacrifice more time than DCP and PR-Net which only require a single forward step for testing, but our model requires much less time than classical iterative methods such as ICP. More randomly selected qualitative results of our model are demonstrated in Figure \ref{fig:4}.

\section{Conclusion}
This paper introduces a novel approach that integrates a learning-based decoder with one optimizable descriptor to our research community for point set registration. With the newly proposed adaptive chamfer distance, our model can be perfectly applied for aligning partial shapes. We conducted experiments on the ModelNet40 datasets to validate the performance of our method.  The results demonstrated that our proposed approach achieved competitive advantages regarding alignment accuracy but sacrifices acceptable computation time in comparison to state-of-the-art approaches. 

\bibliographystyle{IEEEtran}
\bibliography{egbib}
\end{document}